\documentclass[letterpaper]{article} 
\usepackage{aaai24}  
\usepackage{times}  
\usepackage{helvet}  
\usepackage{courier}  
\usepackage[hyphens]{url}  
\usepackage{graphicx} 
\usepackage{natbib}  
\usepackage{caption} 
\usepackage{algorithm}
\usepackage{algorithmic}
\usepackage{tabularx}
\usepackage{amsmath}
\usepackage{amsfonts}       
\usepackage{booktabs}       
\usepackage{multirow}

\usepackage{newfloat}
\usepackage{listings}
\DeclareCaptionStyle{ruled}{labelfont=normalfont,labelsep=colon,strut=off} 
\floatstyle{ruled}
\newfloat{listing}{tb}{lst}{}
\floatname{listing}{Listing}
\title{3D Visibility-aware Generalizable Neural Radiance Fields for Interacting Hands}
\author{
    Xuan Huang\textsuperscript{\rm 1}\equalcontrib, Hanhui Li\textsuperscript{\rm 1}\equalcontrib, Zejun Yang\textsuperscript{\rm 2}, Zhisheng Wang\textsuperscript{\rm 2}, Xiaodan Liang\textsuperscript{\rm 1, \rm 3}\thanks{Corresponding author: xdliang328@gmail.com.}
}
\affiliations{
    \textsuperscript{\rm 1}Shenzhen Campus of Sun Yat-sen University, Shenzhen, China\\
    \textsuperscript{\rm 2}Tencent, Shenzhen, China\\
    \textsuperscript{\rm 3}DarkMatter AI Research, Guangzhou, China\\
}

\begin{document}

\maketitle

\begin{abstract}
Neural radiance fields (NeRFs) are promising 3D representations for scenes, objects, and humans. However, most existing methods require multi-view inputs and per-scene training, which limits their real-life applications. Moreover, current methods focus on single-subject cases, leaving scenes of interacting hands that involve severe inter-hand occlusions and challenging view variations remain unsolved. To tackle these issues, this paper proposes a generalizable visibility-aware NeRF (VA-NeRF) framework for interacting hands. Specifically, given an image of interacting hands as input, our VA-NeRF first obtains a mesh-based representation of hands and extracts their corresponding geometric and textural features. Subsequently, a feature fusion module that exploits the visibility of query points and mesh vertices is introduced to adaptively merge features of both hands, enabling the recovery of features in unseen areas. Additionally, our VA-NeRF is optimized together with a novel discriminator within an adversarial learning paradigm. In contrast to conventional discriminators that predict a single real/fake label for the synthesized image, the proposed discriminator generates a pixel-wise visibility map, providing fine-grained supervision for unseen areas and encouraging the VA-NeRF to improve the visual quality of synthesized images. Experiments on the Interhand2.6M dataset demonstrate that our proposed VA-NeRF outperforms conventional NeRFs significantly. Project Page: \url{https://github.com/XuanHuang0/VANeRF}.
\end{abstract}

\section{Introduction}

\begin{figure*}[t]
  \centering
  \includegraphics[width=1.0\hsize]{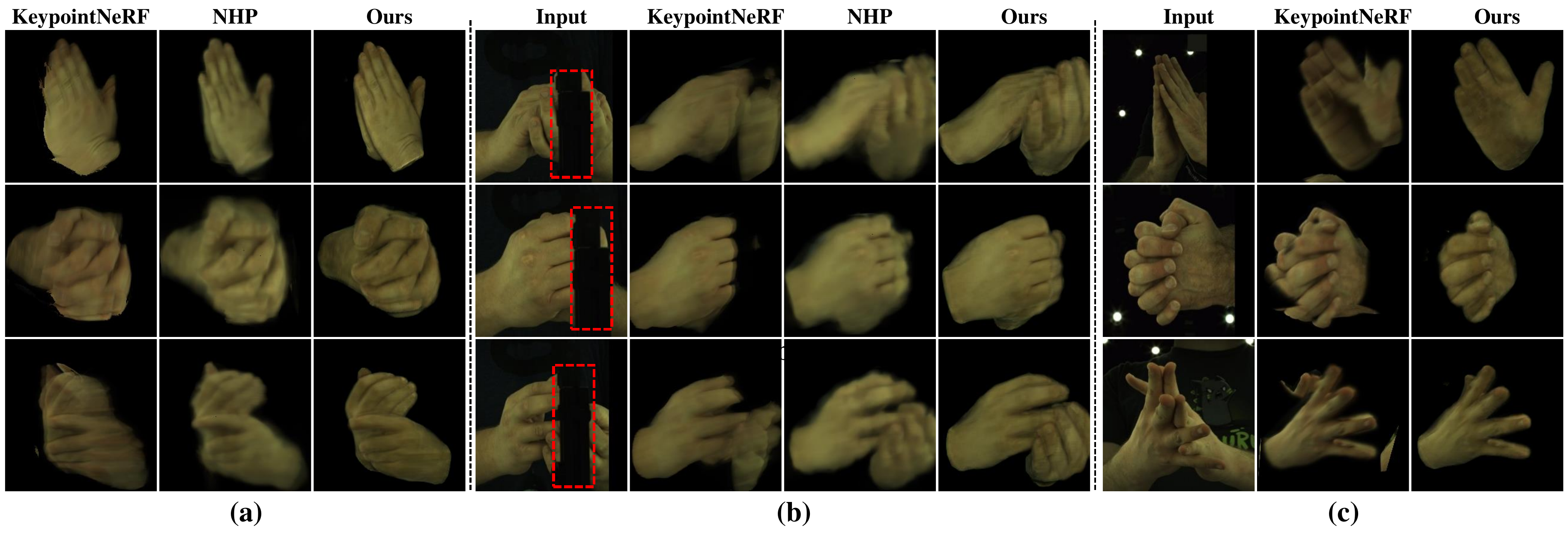}
  \caption{Compared with previous generalizable NeRFs, our visibility-aware NeRF not only (a) generates images of better quality, but also tackles challenging tasks such as (b) inpainting obstructed areas and (c) removing hands in interacting scenes.}
  \label{fig:banner}
\end{figure*}

Recent progress in neural radiance fields (NeRFs) \cite{mildenhall2021nerf,gao2022nerf,niemeyer2022regnerf, Johari_2022_CVPR} is promising, as the continuous implicit representation of NeRFs can be disentangled with spatial volume resolution and generate high-fidelity results. This facilitates interesting research such as human avatars \cite{jiang2022instantavatar}, text-to-3d generation \cite{poole2022dreamfusion}, and large-scale 3D urban scene modeling \cite{turki2022mega}.

However, to synthesize high-quality images, current NeRFs require dozens of well-calibrated multi-view images and hours of per-scene optimization. Although several methods \cite{muller2022instant,chen2022tensorf} have been proposed to reduce the computational cost, it is still expensive and difficult to apply NeRFs in real-life applications where only single-view inputs are available.

Generalizable NeRFs \cite{mihajlovic2022keypointnerf,kwon2021neural} that represent geometry and textures separately seem to be a potential solution for the above issues. Nevertheless, these methods are designed for human bodies and cannot be applied to interacting hands directly. This is because, unlike single-subject human avatars, interacting hands involve challenging factors like severe self/inter-hand occlusions and large view variations \cite{park2022handoccnet,deng2022recurrent}. These factors make it difficult to exploit reliable features and cause artifacts that cannot be ignored.

Therefore, to fulfill application needs and tackle the above challenging factors, this paper aims to design a single-image generalizable NeRF model for interacting hands. In our early exploration, we find that \cite{mihajlovic2022keypointnerf} retains textures and details well but is sensitive to view variations and occlusions, since it adopts the pixel-aligned feature representation \cite{saito2019pifu}. \cite{kwon2021neural} uses a global feature representation that maintains the overall hand structures but is hard to generate fine-grained textures. This indicates that self-adaptive and robust features are the key to constructing feasible NeRFs for interacting hands.

Particularly, we propose a visibility-aware NeRF framework (denoted as VA-NeRF), of which the core is to leverage the visibility of 3D points. Such an idea is natural, because if a 3D point is visible, then its corresponding feature is more reliable. Otherwise, we should select other related points or global features for reference and information complement. Formally, the proposed method achieves this via a visibility-aware feature fusion module that determines the feature of a 3D query point not only by its visibility, but also by vertices selected from hand meshes and by global features. In this way, we can overcome the limitations of feature representation in previous methods.

Moreover, we also propose a visibility-guided adversarial learning strategy to further enhance the synthesized images of VA-NeRF. This is motivated by our observation that the quality of invisible areas in synthesized images is usually worse than that of visible areas. Hence, we propose to encourage the NeRF model to refine invisible areas. However, conventional binary-class discriminators can only provide global supervision by classifying an image as real or fake. Therefore, we design a discriminator that predicts pixel-wise conditional visibility maps, so that it can provide localized and fine-grained supervision for our NeRF model. 

With the above feature fusion module and adversarial learning strategy, the proposed VA-NeRF can synthesize images of interacting hands effectively, and achieve state-of-the-art performance that is validated by experiments on the Interhand2.6M dataset \cite{moon2020interhand2}. In addition, as shown in Figure \ref{fig:banner}, our VA-NeRF can accomplish tasks that are challenging for conventional methods, such as removing hands and recovering invisible areas, and consequently benefit downstream applications like hand pose estimation \cite{meng20223d}.

In summary, the contributions of this paper can be listed as follows:

$\bullet$ To the best of our knowledge, this paper proposes the first single-image generalizable neural radiance field model for interacting hands.

$\bullet$ A visibility-aware feature fusion module is proposed, which adaptively leverages various visual features (global features, pixel-wise aligned features, and symmetric hand features) to tackle challenging occlusions and view variations.

$\bullet$ An adversarial learning strategy guided by visibility maps is introduced, which further improves the visual quality of synthesized two-hand images.

\section{Related Work}

\textbf{Neural radiance fields}.
In recent years, NeRF \cite{mildenhall2021nerf} has been widely studied in the area of 3D human reconstruction due to its stunning results. Many efforts have been made to adapt NeRFs to high-fidelity novel-view synthesis of human performers/avatars \cite{deng2022depth,mildenhall2022nerf,johari2022geonerf,niemeyer2022regnerf, Johari_2022_CVPR,martin2021nerf}. \cite{raj2021pixel,wang2021ibrnet,yu2021pixelnerf} propose to utilize pixel-aligned features to learn generalized models from sparse views. Besides, KeypointNeRF \cite{mihajlovic2022keypointnerf} proposes to encode relative spatial 3D information with sparse 3D key points as references. Recent approaches \cite{peng2021neural,kwon2021neural} have incorporated parametric hand meshes as geometry priors to reduce the dependence on multi-view captures. NHP \cite{kwon2021neural} applies a temporal transformer and a multi-view transformer to fuse visual features conditioned on SMPL \cite{loper2015smpl}. For a smoother surface prediction and better 3D consistency, \cite{or2022stylesdf,hong2022eva3d,corona2022lisa} merge implicit representations into the density prediction procedure of NeRF. Especially, \cite{corona2022lisa} is designed for single-hand images. A concurrent approach that aims at NeRF for hands is proposed in \cite{guo2023handnerf}, yet it requires multi-view inputs while we focus on monocular single-image scenes. SHERF \cite{hu2023sherf} recovers animatable 3D humans from a single input image by extracting 3D-aware hierarchical features, including global, point-level, and pixel-aligned features, to facilitate informative encoding. The proposed method differs from SHERF in the aspect of tasks and feature fusion modules. SHERF combines features via self-attention while our module is conditioned on visibility.

Unlike previous methods, our VA-NeRF is designed for single-view generalizable interacting-hand image synthesis. Moreover, conventional generalizable methods rely on the single local/global representation and hence suffer from occlusions and view variations in our task. On the contrary, our VA-NeRF exploits visibility in both feature fusion and adversarial learning, which helps to improve the visual quality of synthesized images significantly.

\begin{figure*}[t]
  \centering
  \includegraphics[width=1.0\hsize]{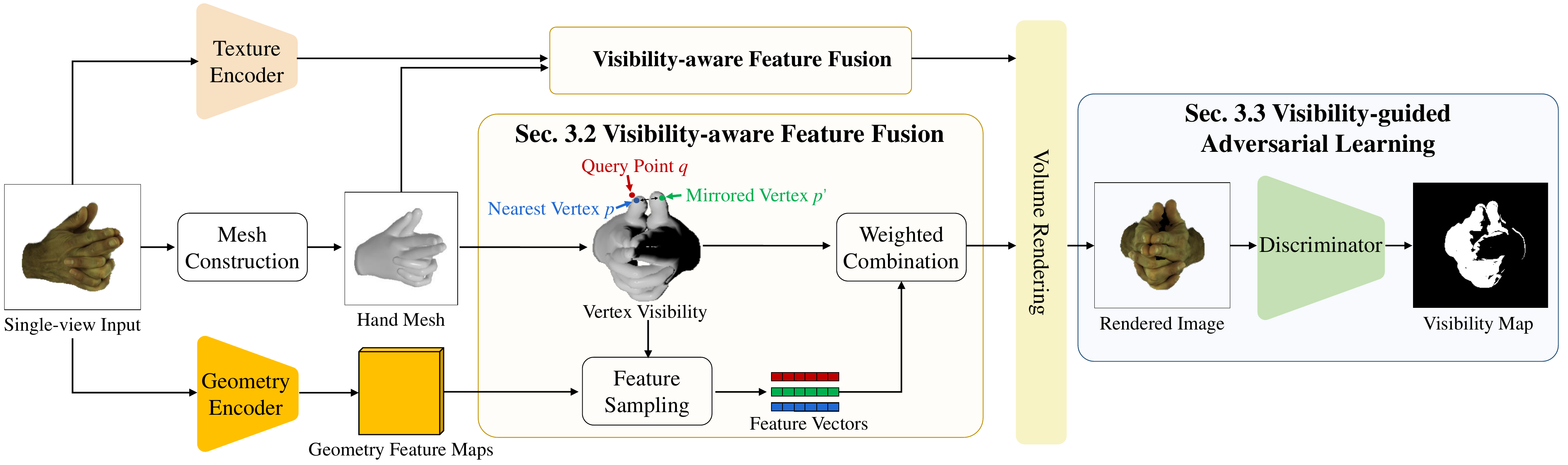}
  \caption{The framework of VA-NeRF. It consists of two key components and both of them are designed to leverage the visibility of 3D points. The first one is the visibility-aware feature fusion module that estimates appropriate features for query points, while the second one is the visibility-guided adversarial learning strategy that is used to enhance synthesized results.}
  \label{fig:framework}
\end{figure*}

\noindent\textbf{3D hand reconstruction}.
Parametric hand models such as MANO \cite{romero2022embodied} have enabled hand mesh to be reconstructed via inferring a set of pose and shape parameters. Prior approaches belonging to this type \cite{zhang2021interacting,chen2021camera,kulon2020weakly,zhou2020monocular} can achieve parameter fitting on single images. Moreover, the MANO-HD model \cite{chen2022hand} is developed as a high-resolution mesh topology to fit personalized hand shapes and generate smooth geometry. Implicit representations \cite{chen2022alignsdf,karunratanakul2021skeleton}, such as signed distance fields, have received considerable attention recently, as theoretically, they can approximate any geometry details. A dataset that consists of diverse textures and hand accessories is introduced in \cite{gaodart2022}. 

\noindent\textbf{3D-aware adversarial learning}.
Recent years have witnessed the great success of generative adversarial networks (GANs) \cite{goodfellow2020generative} in photorealistic image generation. While GANs based on 2D latent space lack 3D understanding, 3D generative models \cite{schwarz2020graf,chan2021pi,or2022stylesdf} enable explicit camera control and render more realistic images from random viewpoints. \cite{hong2022eva3d,deng2022gram,niemeyer2021giraffe,xu2021generative} combine implicit NeRF with GAN for better view-consistency and more detailed 3d shape. To reduce the computational cost, EG3D \cite{chan2022efficient} proposes an efficient tri-plane structure while EVA3D \cite{hong2022eva3d} divides the human body into local parts and omits unnecessary computations in blank space.

\section{Methodology}
The goal of this paper is to construct a single-view generalizable NeRF for interacting hands and to tackle occlusions and view variations. To this end, we propose a visibility-aware NeRF framework that leverages the visibility of query points and hand mesh vertices through a feature fusion module and an adversarial learning strategy. These two approaches serve distinct purposes: the former aims to infer complementary features for occluded regions, while the latter seeks to enhance the quality of rendered results.

\subsection{VA-NeRF Framework}
\label{sec:framework}

\begin{figure*}[t]
  \centering
  \includegraphics[width=1.0\hsize]{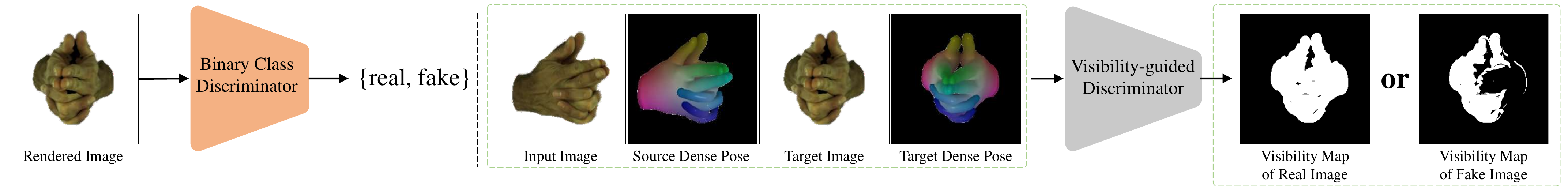}
  \caption{Comparison between the traditional binary-class discriminator (left) and the proposed visibility-guided discriminator (right). Note that the visibility map is conditioned on the input view and whether the target image is real or synthesized.}
  \label{fig:dis}
\end{figure*}

\textbf{Problem formulation}. Our task is to construct a generalizable NeRF that can render novel views of an arbitrary pair of interacting hands given a single input image. Specifically, the NeRF can be formulated as a function $f:(q,d,I)  \to (c, \sigma)$, where $q \in \mathbb{R}^3$ is a query point, $d \in \mathbb{R}^3$ denotes a viewing direction, and $I$ is the input image. The output $c \in \mathbb{R}^3$ and $\sigma \in \mathbb{R}$ are the color and volume density of the query point, respectively. With query points densely sampled in a 3D volume, we infer their colors and densities and synthesize a target-view image by volume rendering \cite{mildenhall2021nerf}.

To complete the above task, we propose the VA-NeRF framework as shown in Fig. \ref{fig:framework}. As the core of VA-NeRF is to leverage the visibility of query points and mesh vertices, we first construct hand meshes by fitting the MANO parametric model \cite{romero2017embodied} to the input image. Following \cite{mihajlovic2022keypointnerf}, we employ two encoder branches to disentangle geometric and textual features: an hourglass network \cite{newell2016stacked} serves as the geometry encoder and a convolutional neural network (CNN) with residual connections \cite{johnson2016perceptual} serves as the texture encoder. Subsequently, we introduce a visibility-aware feature fusion (VAFF) module in each feature branch to obtain the visibility-enhanced feature of the query point. We utilize a multilayer perception (MLP) to infer the color $c$ from the texture feature. For the density $\sigma$, we follow \cite{hong2022eva3d} to estimate a deviated signed distance field (SDF) with respect to the hand meshes:
\begin{equation}
    \sigma(q) = {w^{ - 1}}\mathrm{sig}( - (s(q) + \delta (q))/w),
    \label{eq:sdf}
\end{equation}
where $w \in \mathbb{R}$ is a weighting parameter optimized along with the network. $\mathrm{sig}(\cdot)$ represents the sigmoid function. $s(q) \in \mathbb{R}$ is the explicit SDF value of the query point calculated with the mesh surfaces as the zero level-set, while $\delta (q) \in \mathbb{R}$ is the deviation inferred by another MLP that takes the geometry feature of $q$ as input. The target-view image is then generated by using an off-the-shelf differentiable renderer \cite{wang2021ibrnet}.

In addition, our VA-NeRF network is optimized using an adversarial learning strategy that exploits visibility to provide additional supervision. Our adversarial learning strategy relies on a discriminator that learns not only to distinguish between real and synthesized images but also to predict pixel-wise visibility maps conditioned on input and target views. Correspondingly, through the competition between the VA-NeRF network and the discriminator, we not only require the VA-NeRF network to synthesize high-fidelity images, but also encourage it to improve the quality of invisible areas in results (in order to ``deceive" the discriminator). Details of our network architecture are available in the supplemental material on our project page.

\subsection{Visibility-aware Feature Fusion}
\label{sec:attn}

Traditional approaches that rely solely on either pixel-aligned or global features are ineffective in addressing the challenges caused by heavy occlusions and complex interacting poses, as these features are hard to remain reliable under such circumstances. To overcome this limitation, we introduce the VAFF module, which adaptively selects and combines a set of features for each query point.

To elaborate, we take the VAFF module in the texture feature branch as an example. For the sake of presentation conciseness, we assume that the dimensions of all features for fusion are $D$. Given a query point $q$, we locate its nearest neighboring mesh vertex $p$ on one hand. We assume that both hands have the same topology (i.e., vertices are ordered following the MANO model) and consider the vertex with the same order on the other hand as the mirrored point $p'$. Note that the above topological assumption does not impose the constraint that both hands are of the same pose. $q$, $p$, and $p'$ are projected onto the 2D image plane, and their corresponding feature vectors are retrieved via bilinear interpolation on the texture feature maps. Let $k(q), m(p), n(p') \in \mathbb{R}^D$ denote the retrieved feature vectors of $q$, $p$, and $p'$, respectively. We consider the following weighted concatenation $t(q) \in \mathbb{R}^{6D}$ as the enhanced feature of $q$:
\begin{equation}
\begin{split}
        t(q) = [{a_\varphi }\varphi (q), \: {a_k}k(q), \: {a_m}m(p), \\ {a_n}n(p'), \: a_g^l{{\bf{g}}^l}, \: a_g^r{{\bf{g}}^r}],
\end{split}
\end{equation}
where ${a_\varphi }, {a_k}, {a_m}, {a_n}, a_g^l, a_g^r \in [0,1]$ are feature weights. $\varphi (q) \in \mathbb{R}^D$ is the spatial feature of $q$ obtained by positional encoding. ${{\bf{g}}^l}, {{\bf{g}}^r} \in \mathbb{R}^D$ are the global average texture features of the left hand and right hand. 

To ensure that $t(q)$ can select and fuse appropriate features, we define a weighting function governed by 3D point visibility. Particularly, let $a \in \{{a_\varphi }, {a_k}, {a_m}, {a_n}, a_g^l, a_g^r\}$ be an arbitrary weight and $v(p,d) \in \{0, 1\}$ denote the visibility of point $p$ from the viewing direction $d$. $v(p,d) = 1$ if $p$ is visible, otherwise $v(p,d) = 0$. We calculate $a$ via a function $\mu: \mathbb{R}^{6D+3} \to [0, 1]$ as follows:
\begin{equation}
\begin{split}
        a = \mu(v(q,d), \: v(p,d), \: v(p',d), \: \varphi (q), \\ k(q), \: m(p), \: n(p'), \: {{\bf{g}}^l}, \: {{\bf{g}}^r}).
\end{split}
    \label{eq:feature}
\end{equation}
We implement $\mu$ simply by a MLP. The architecture of the VAFF module in the geometry feature branch is similar, except that it does not contain the global average of geometry feature maps (since geometry information is rather local).

\textbf{Discussion}. Our VAFF is based on the assumption that the feature vector of a visible point should be assigned a high weight. Additionally, we also consider $p$ because spatially closed points tend to share similar features. As for $p'$, it is included due to the structural symmetry of human hands. $p$ and $p'$ together facilitate the exploitation of both local and long-range feature dependencies, so to tackle pose and view variations. In cases where $q$, $p$, and $p'$ are all invisible, the global average feature vectors can still serve as a coarse approximation to the texture feature of $q$. This is feasible as most areas of the hands have similar textures. Consequently, the VAFF module effectively leverages the strengths of both local pixel-aligned features for detail preservation and global features for hand structure preservation.

\begin{figure*}[t]
  \centering
  \includegraphics[width=1.0\hsize]{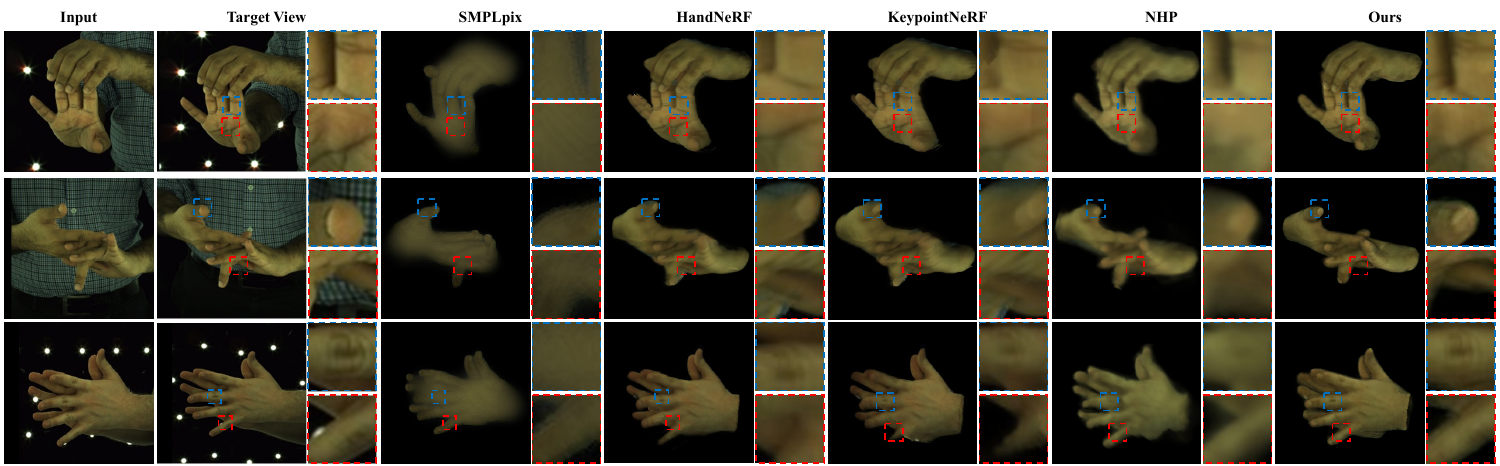}
  \caption{Visual comparison of the proposed method against state-of-the-art methods. Results of the proposed method better preserve hand structures and textures.}
  \label{fig:quality}
\end{figure*}

\subsection{Visibility-guided Adversarial Learning}
\label{sec:dis}

Due to the lack of information, the visual quality of invisible areas in synthesized images tends to be lower than that of visible areas. Therefore, here we propose the visibility-guided adversarial learning (VGAL) strategy to facilitate quality improvement in invisible areas.

To begin with, a discriminator that can identify invisible areas is necessary. This can be achieved by conducting hand mesh rasterization to generate ground-truth visibility maps for supervised learning. Specifically, given an input-view image $I$ and a target-view image $I_t$, where $I$ and $I_t$ are of size $H \times W$, the discriminator can be formulated as $\Phi(I, I_t): \mathbb{R}^{H\times W} \times \mathbb{R}^{H\times W} \to \{0, 1\}^{H \times W}$. It is important to note that $I_t$ can be either real or synthesized and the ground-truth visibility map $V_t$ should be conditioned on this. We define the following two criteria for generating $V_t$: 

(i) If $I_t$ is real, we consider the foreground (hand areas) of $I_t$ to be visible, 

(ii) otherwise $V_t$ is rendered with vertex visibility computed in the input view. 

An example of $V_t$ is shown in Figure \ref{fig:dis}. With these criteria, the discriminator needs to recognize invisible areas in synthesized images while the VA-NeRF network is encouraged to generate results that have visibility maps similar to those of real target-view images.

We implement $\Phi$ by a CNN with the sigmoid function as its last activation function. Except $I$ and $I_t$, we also generate dense correspondence maps \cite{guler2018densepose} as the auxiliary inputs to $\Phi$, which are used to provide structural priors of human hands. Finally, the objective functions of our VGAL are defined as follows:
\begin{equation}
\mathcal{L}=\lambda_{rgb}\mathcal{L}_{rgb}+\lambda_{VGG}\mathcal{L}_{VGG}+\lambda_{adv}\mathcal{L}_{adv}+\lambda_{vis}\mathcal{L}_{vis},
\label{eq:loss}
\end{equation}
where $\lambda_{rgb}$, $\lambda_{VGG}$, $\lambda_{adv}$ and $\lambda_{vis}$ are user-defined loss weights. $\mathcal{L}_{rgb}$ and $\mathcal{L}_{VGG}$ are the $l1$ loss and the perceptual loss \cite{johnson2016perceptual} between the target image and the synthesized image. $\mathcal{L}_{adv}$ is the non-saturating GAN loss \cite{mescheder2018training,hong2022eva3d} widely used in adversarial learning. $\mathcal{L}_{vis}$ is introduced to supervise visibility learning, which is formulated as the pixel-wise binary cross entropy between the predicted visibility map $V$ and $V_t$ as follows:
\begin{equation}
    \mathcal{L}_{vis}(V, V_{t}) = - (V_{t} \odot \mathrm{log}V + (1-V_{t}) \odot \mathrm{log}(1 - V)),
    \label{eq:vis_loss}
\end{equation}
where $\odot$ denotes the element-wise dot product. We follow the common practice of adversarial learning \cite{mescheder2018training} to optimize the VA-NeRF network and the discriminator alternately with their corresponding loss terms. Take $\mathcal{L}_{vis}$ as an example, during the training phase of the VA-NeRF network, $\mathcal{L}_{vis}$ is calculated with $V_t$ conditioned on the real target image, and its gradients are propagated backward from the discriminator to the VA-NeRF network; while at the training phase of the discriminator, $\mathcal{L}_{vis}$ is calculated with $V_t$ determined by whether the target image is real or rendered.

\section{Experiments}

In this section, we validate the effectiveness of the proposed VA-NeRF method via extensive experiments. Due to the page limitation, interested readers can refer to the supplemental material for more implementation details, experimental results, and discussions.

\subsection{Setup}

\textbf{Dataset}.
Our experiments are conducted on the large-scale Interhand2.6M \cite{moon2020interhand2} dataset that consists of single and interacting hand images with various subjects, poses, and views. As the scope of this paper is to construct NeRFs for interacting hands, we select a subset of images on Interhand2.6M, which contains 143,893 training images and 9,475 test images in total. We further cropped out all hand regions based on the bounding boxes provided by the dataset and resize all images to $256 \times 256$.

\noindent\textbf{Implementation details}.
Our network is implemented using PyTorch and trained with the Adam optimizer \cite{kingma2014adam} with a batch size of 4. For both the VA-NeRF and the discriminator, their initial learning rates are set to $1 \times 10^{-3}$ and decay by half four times (at the 2nd, 5th, 10th, and 20th epoch respectively) during training. The whole training process takes about 40 hours on four NVIDIA RTX 3090 GPUs. Loss weights in Eq. (\ref{eq:loss}) are set as $\lambda_{rgb}=10.0, \lambda_{VGG}=1.0, \lambda_{adv}=0.1, \lambda_{vis}=0.1$. The total number of training epochs is 30. As in \cite{mihajlovic2022keypointnerf}, we adopt a coarse-to-fine rendering strategy during training that first renders patches by accumulating color and density values of 64 sampled points along a camera ray, and then 128 sampled points for fine-grained rendering. 

\noindent\textbf{Baselines}.
As there is no open-source baseline for interacting hands, we adopt two state-of-the-art generalizable NeRFs designed for humans, including NHP \cite{kwon2021neural} and KeypointNeRF \cite{mihajlovic2022keypointnerf}. Besides, although HandNeRF \cite{guo2023handnerf} is non-generalizable, we still combine its core module, i.e., the depth-guided density optimization strategy with KeypointNeRF for comparison. We also choose SMPLpix \cite{prokudin2021smplpix} as an image-space baseline. All networks are trained with the same experimental setting for fair comparison. We select three widely-used evaluation metrics, including peak signal-to-noise ratio (PSNR) \cite{sara2019image}, structural similarity index (SSIM) \cite{wang2004image}, and learned perceptual image patch similarity (LPIPS) \cite{zhang2018unreasonable}. 

\begin{figure}[t]
  \centering
  \includegraphics[width=1.0\hsize]{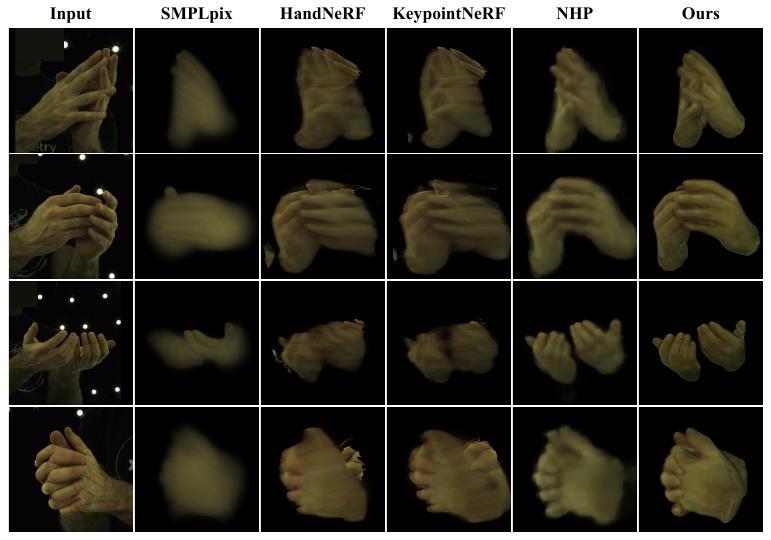}
  \caption{Qualitative examples of novel-view rendering with large view variations (rotation angles $>$ 30 degrees).}
  \label{fig:large_view}
\end{figure}

\begin{figure}[t]
  \centering
  \includegraphics[width=1.0\hsize]{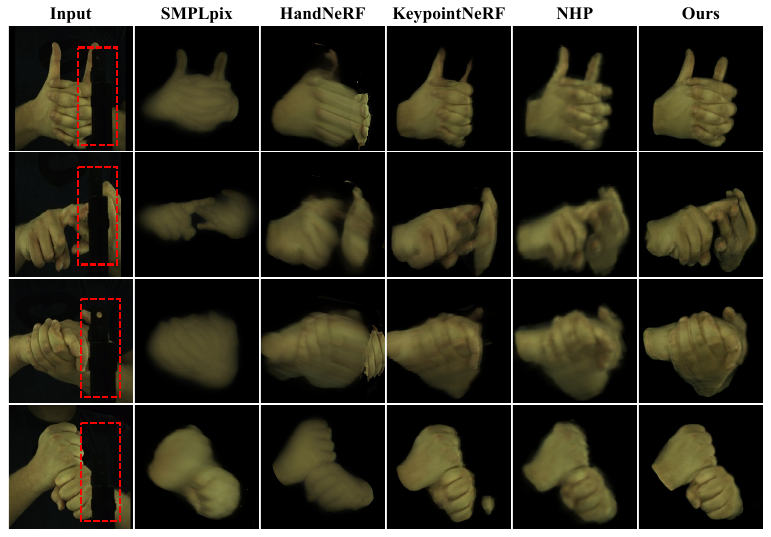}
  \caption{Qualitative comparison of synthesized images in scenes involving severe occlusions.}
  \label{fig:occlusion}
\end{figure}

\begin{table}[t]
    \caption{Comparison with state-of-the-art methods on Interhand2.6M.}
    \label{tab:Comparison}
    \centering
    \begin{tabular}{cccc}
        \toprule
        Method     & PSNR↑    & SSIM↑    & LPIPS↓     \\
        \midrule
        SMPLpix & 22.49 & 0.82 & 0.33 \\
        KeypointNeRF & 23.49   & 0.82  & 0.27   \\
        NHP & 23.63   & 0.83  & 0.33     \\ 
        HandNeRF & 23.68 & 0.83 & 0.27 \\ \hline
        Ours     &\textbf{25.01}   & \textbf{0.86}   & \textbf{0.21}  \\
        \bottomrule
    \end{tabular}
\end{table}

\begin{table}[t]
    \caption{Performance comparison among generalizable NeRFs under large view variations ($>$ 30 degrees).}
    \label{tab:quanti_view}
    \centering
    \begin{tabular}{cccc}
    \toprule
        Method     & PSNR↑    & SSIM↑    & LPIPS↓     \\
        \midrule
        KeypointNeRF & 22.35 & 0.77 & 0.34 \\
        NHP & 22.98 & 0.80 & 0.36 \\
        Ours & \textbf{24.23} & \textbf{0.84} & \textbf{0.22} \\
        \bottomrule
    \end{tabular}
\end{table}

\begin{table}[!t]
\centering
\caption{Performance comparison among generalizable NeRFs under occlusions.}
\label{tab:quanti_mask}
\small
\begin{tabular}{ccccc}
\toprule
Method  & Mask Ratio & PSNR↑    & SSIM↑    & LPIPS↓     \\
\midrule
KeypointNeRF & \multirow[c]{3}{*}{0.1} &  24.23  &  0.84  &  0.23  \\
NHP &  &  23.67   &  0.83   &  0.32     \\
Ours &  &  \textbf{25.61}    &  \textbf{0.86}     &  \textbf{0.19}      \\ \hline
KeypointNeRF & \multirow[c]{3}{*}{0.2} &  22.88  &   0.81   & 0.25      \\
NHP &  & 23.60 &  0.83   &     0.33   \\
Ours &  &   \textbf{25.50}   &  \textbf{0.86}    &   \textbf{0.19}   \\ \hline
KeypointNeRF & \multirow[c]{3}{*}{0.3} &  21.38  &   0.78   &     0.29  \\
NHP &  &  23.50   &   0.82  &   0.33    \\
Ours &  &   \textbf{24.93}    &  \textbf{0.85}     &  \textbf{0.21}  \\
\bottomrule
\end{tabular}
\end{table}

\begin{figure}[t]
  \centering
  \includegraphics[width=1.0\hsize]{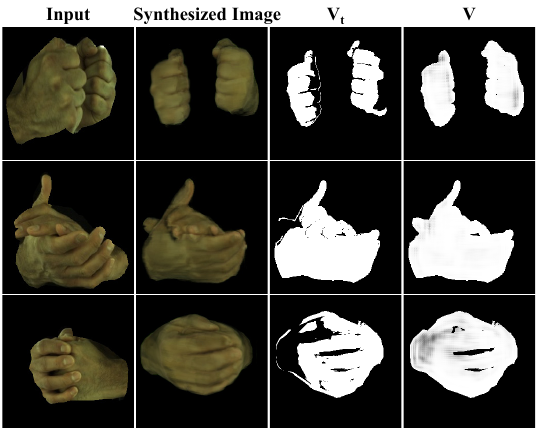}
  \caption{Visualization of visibility maps. $V_t /V$ are target-view/predicted visibility maps. Our NeRF successfully fools the discriminator as most areas in its synthesized images are recognized as visible.}
  \label{fig:vis_dis}
\end{figure}

\begin{table}[t]
    \caption{Ablation study on selected features.}
    \label{tab:feats}
    \centering
    \begin{tabular}{cccc}
        \toprule
        Feature     & PSNR↑    & SSIM↑    & LPIPS↓     \\
        \midrule
        $q$ & 24.86 & 0.87 & 0.18 \\
        $q + p$ & 25.06 & 0.86 & 0.19 \\
        $q + p'$ & 25.47 & 0.86  & 0.18   \\
        $q + p + p'$ & \textbf{25.74} & \textbf{0.87}  & \textbf{0.18}     \\
        \bottomrule
    \end{tabular}
\end{table}

\begin{table}[t]
    \caption{Ablation study on attention strategies.}
    \label{tab:attn}
    \centering
    \begin{tabular}{cccc}
    \toprule
        Method     & PSNR↑    & SSIM↑    & LPIPS↓     \\
        \midrule
        KeyPointNeRF & 24.37 & 0.85 & 0.24 \\ \hline
        Attn. w/o Vis. & 24.74 & 0.86 & 0.20 \\
        Attn. w/ Vis. & \textbf{25.74} & \textbf{0.87} & \textbf{0.18} \\ 
        \bottomrule
    \end{tabular}
\end{table}

\begin{table}[t]
\centering
\caption{Ablation study on discriminators.}
\label{tab:disc}
\begin{tabular}{ccccc}
\toprule
Scene  & Method & PSNR↑    & SSIM↑    & LPIPS↓     \\
\midrule
View & Bi. Dis. &  23.77  &  0.83  &  0.24  \\ \cline{2-5}
Variation & Vis. Dis.  &  \textbf{24.34}  &  \textbf{0.84}   &  \textbf{0.21}     \\ \hline
\multirow[c]{2}{*}{Occlusion} & Bi. Dis. &  24.57  &  0.85  &  0.20  \\ \cline{2-5}
 & Vis. Dis.  &  \textbf{25.43}   &  \textbf{0.86}  &  \textbf{0.19}    \\
\bottomrule
\end{tabular}
\end{table}

\subsection{Comparison with State-of-the-arts}

\textbf{Quantitative comparison}. Table \ref{tab:Comparison} reports the quantitative results of our VA-NeRF against the baselines on Interhand2.6M. We can see that VA-NeRF raises the PSNR and the SSIM of state-of-the-art methods from 22.49 to 25.01 and 0.82 to 0.86, and also reduces the LPIPS from 0.33 to 0.21. These performance gains indicate that the synthesized results of VA-NeRF better preserve the hand structures and details (PSNR and SSIM) and are more realistic (LPIPS).

\noindent\textbf{Qualitative comparison}. Figure \ref{fig:quality} provides the visual comparison between our VA-NeRF and the baselines. Compared with the baselines, the results of VA-NeRF are of better quality and have fewer artifacts. SMPLpix relies on image-space transfer only and hence its results are over-smoothed. The positional encoding in KeypointNeRF relies on MANO joints only and hence it is affected by joint estimation errors severely. HandNeRF alleviates depth ambiguities but is still hard to maintain hand shapes under complex interacting cases. NHP adopts global features and hence the hand structures are preserved well in its results. However, it omits details like wrinkles and nails. Our method adopts the VAFF module to select and merge features, hence it alleviates the disadvantages of baselines successfully. 

Moreover, since the major scope of this paper is to tackle large view variations and heavy occlusions, we also evaluate the proposed method in these two cases. 

\noindent\textbf{Robustness to large view variations}. Novel-view synthesis with large view variations (rotation angles $>$ 30 degrees) requires NeRFs to model long-range feature dependencies. Figure \ref{fig:large_view} shows the results of the proposed VA-NeRF against baselines in this case. We can see that the results of baselines are severely blurred. On the contrary, the results of our VA-NeRF are still satisfying. We also provide the quantitative comparison between VA-NeRF and the two generalizable baselines in Table \ref{tab:quanti_view}. Thanks to the VAFF module, our VA-NeRF can model the long-range dependencies among symmetric mesh vertices, and combine global features adaptively to address large view variations. 

\noindent\textbf{Robustness to heavy occlusions}. The task of image synthesis under heavy occlusions is difficult as well, since certain regions may remain obscured from multiple viewing angles. Figure \ref{fig:occlusion} demonstrates visual examples of VA-NeRF in this case. It is clear that the proposed method successfully recovers unseen areas with realistic textures and structures, while some results of the baselines are distorted and details are not preserved well. Moreover, we generate occluded test images by adding masks centered at images with different ratios (i.e., 0.1, 0.2, and 0.3 of the image size) to conduct quantitative analysis. From the results in Table \ref{tab:quanti_mask}, we can see that our VA-NeRF better maintains its performance compared with the baselines under occlusions.

\subsection{Ablation Study}

\noindent\textbf{Feature selection}. To verify that each selected feature ($q$, $p$, and $p'$) in our VA-NeRF is necessary, we evaluate the performance of all possible feature combinations. The results are shown in Table \ref{tab:feats} and they reflect that each feature does bring performance gains. The best performance is obtained by using all three features.

\noindent\textbf{Effectiveness of VAFF}. To validate the effectiveness of visibility in our feature fusion module, we implement a variant that learns the weights of features without the guidance of visibility. Table \ref{tab:attn} reports the performance of VAFF and the variant (denoted as Attn. w/o Vis.). We can see that, with the help of visibility, our feature fusion module obtains significantly better performance.

\noindent\textbf{Effectiveness of VGAL}. We also evaluate the effectiveness of visibility maps in adversarial learning. We implement a conventional binary discriminator (denoted as Bi. Dis.) and compare it with our visibility-guided discriminator (Vis. Dis.). The results are shown in Table \ref{tab:disc} and it is clear that the proposed discriminator outperforms the binary one by large margins in scenes with view variations and occlusions. 

\noindent\textbf{Visualization of visibility maps}. The predictions of the proposed discriminator are shown in Figure \ref{fig:vis_dis}. We can see that the VA-NeRF network does learn to compete with the discriminator effectively, as most regions are recognized as visible by the discriminator. Hence, the proposed VAGL strategy has achieved our goals successfully.

\section{Conclusion}
In this paper, we introduce a single-image generalizable visibility-aware neural radiance field framework for image synthesis of interacting hands. The proposed framework leverages the visibility of 3D points for feature fusion and adversarial learning. Our feature fusion is achieved by fusing features of reference vertices closely related to query points, with fusion weights determined by point visibility. Our adversarial learning is accomplished through the training of a pixel-wise discriminator capable of estimating visibility maps. With these two components cooperating together, the proposed method can obtain reliable features and high-quality results, even in challenging scenarios involving heavy occlusions and large view variations. The proposed method is evaluated on Interhand2.6M  and obtains performance superior to state-of-the-art generalizable models.

\section{Acknowledgments} 
This work was supported in part by National Key R\&D Program of China under Grant No. 2020AAA0109700,  Guangdong Outstanding Youth Fund (Grant No. 2021B1515020061), National Natural Science Foundation of China (NSFC) under Grant No. 61976233, No. 92270122, No. 62372482 and No. 61936002, Mobility Grant Award under Grant No. M-0461, Shenzhen Science and Technology Program (Grant No. RCYX20200714114642083), Shenzhen Science and Technology Program (Grant No. GJHZ20220913142600001), Nansha Key R\&D Program under Grant No.2022ZD014 and Sun Yat-sen University under Grant No. 22lgqb38 and 76160-12220011. 

\small
\bibliography{mybib}

\end{document}